\documentclass[5p,times,procedia]{elsarticle}
\usepackage{ecrc}
\usepackage{multirow}
\usepackage{booktabs}
\usepackage{amsmath}

\volume{00}

\firstpage{1}

\journalname{Procedia CIRP}

\runauth{Author name}

\jid{trpro}

\usepackage{amssymb}
\usepackage[figuresright]{rotating}
\usepackage{subfigure}
\usepackage[bookmarks=false]{hyperref}
    \hypersetup{colorlinks,
      linkcolor=blue,
      citecolor=blue,
      urlcolor=blue}

\begin{document}
\begin{frontmatter}

\dochead{31st CIRP Conference on Life Cycle Engineering (LCE 2024)}%

\title{6D Pose Estimation on Point Cloud Data through Prior Knowledge Integration: A Case Study in Autonomous Disassembly}

\author[a]{Chengzhi Wu *$^\dag$}
\author[a]{Hao Fu $^\dag$}
\author[b]{Jan-Philipp Kaiser}
\author[c]{Erik Tabuchi Barczak}
\author[d]{\\Julius Pfrommer}
\author[b]{Gisela Lanza}
\author[c]{Michael Heizmann}
\author[a,d]{Jürgen Beyerer}

\address[a]{Institute for Anthropomatics and Robotics, Karlsruhe Institute of Technology, Kaiserstraße 12, 76131 Karlsruhe, Germany}
\address[b]{wbk Institute of Production Science, Karlsruhe Institute of Technology, Kaiserstraße 12, 76131 Karlsruhe, Germany}
\address[c]{Institute of Industrial Information Technology, Karlsruhe Institute of Technology, Hertzstraße 16, 76187 Karlsruhe, Germany}
\address[d]{Fraunhofer Institute of Optronics, System Technologies and Image Exploitation IOSB, Fraunhoferstraße 1, 76131 Karlsruhe, Germany}

\aucores{* Corresponding author. Tel.: +49-(0)1523 8476995. {\it E-mail address:} chengzhi.wu@kit.edu \hspace{2cm} $^\dag$ Equal Contribution}

\begin{abstract}
The accurate estimation of 6D pose remains a challenging task within the computer vision domain, even when utilizing 3D point cloud data. Conversely, in the manufacturing domain, instances arise where leveraging prior knowledge can yield advancements in this endeavor. This study focuses on the disassembly of starter motors to augment the engineering of product life cycles. A pivotal objective in this context involves the identification and 6D pose estimation of bolts affixed to the motors, facilitating automated disassembly within the manufacturing workflow. Complicating matters, the presence of occlusions and the limitations of single-view data acquisition, notably when motors are placed in a clamping system, obscure certain portions and render some bolts imperceptible. Consequently, the development of a comprehensive pipeline capable of acquiring complete bolt information is imperative to avoid oversight in bolt detection. In this paper, employing the task of bolt detection within the scope of our project as a pertinent use case, we introduce a meticulously devised pipeline. This multi-stage pipeline effectively captures the 6D information with regard to all bolts on the motor, thereby showcasing the effective utilization of prior knowledge in handling this challenging task. The proposed methodology not only contributes to the field of 6D pose estimation but also underscores the viability of integrating domain-specific insights to tackle complex problems in manufacturing and automation.
\end{abstract}


\end{frontmatter}

\section{Introduction}
\label{sec:intro}
In today's context, within the framework of life cycle engineering (LCE), the importance of End-of-Life (EOL) strategies has been heightened due to growing environmental concerns. As one of the EOL strategies, remanufacturing salvages the efforts that were initially used to shape the product and presents opportunities for creating further business profits \cite{sundin2004product}.
In remanufacturing, the used products are reintegrated into the production line as a practical application of LCE. After disassembly and additional processing, the individual parts are reassembled into remanufactured products, which should possess functionality and quality equivalent to that of a new product \cite{barquet2013integrated}.
In contrast to traditional manufacturing, remanufacturing is challenging due to the complexity, uncertainty, and inconsistency of the returned products \cite{rizova2020systematic}.
Consequently, the primary processes in the current remanufacturing are still carried out manually, which limits its widespread adoption \cite{kurilova2018remanufacturing}.
In order to automate this process, a flexible and adaptive system is required.

\begin{figure*}[t]
\centering
\includegraphics[width=0.95\linewidth, trim=0 20 0 10, clip]{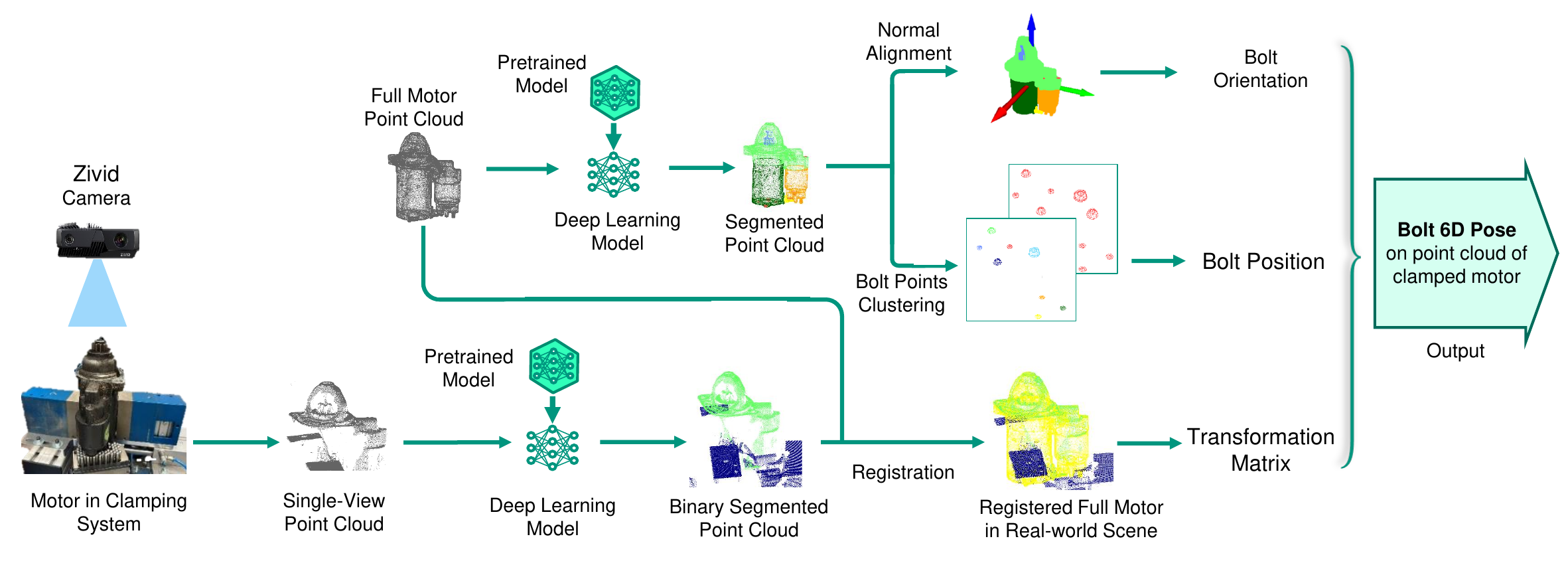}
\caption{Full pipeline of our proposed method. A starter motor is clamped in the clamping system and scanned by the Zivid camera. Extra input data of the full motor point cloud provides the prior knowledge of the motor model. After registration, the 6D pose information of bolts under real-world scenes can be directly given.}
\label{fig:pipeline}
\end{figure*}

The AgiProbot project \cite{lanza2022agiles, wu2023sim2real, Wu2023SynMotorAB} focuses on such automated disassembly system of end-of-life electric motors. 		
As shown in the leftmost part of Fig. \ref{fig:pipeline}, the processed starter motor is clamped and disassembled by a multifunctional robotic arm.	
Concerning the bolts on the motor, besides part segmentation, accurate 6D pose, which encompasses their 3D position and 3D orientation, 
is essential to facilitate the robotic arm's spatial movement to the correct position and ensure its disassembly movement in the correct direction.		
However, due to cost and space constraints, the 3D camera is usually installed in a fixed position, which can provide only single-view point clouds. Meanwhile, under the rigidity and stability requirements of the clamping system, the motor remains in a random but immovable position during the disassembly process. This results in low-quality scanning or even view obstruction of numerous critical features, such as bolts on the complex motor.
In this paper, we propose an ingenious method that uses extra prior knowledge to compute precise bolt 6D pose, even under these unfavorable conditions.

Driven by the demands and product iterations, the starter motors exhibit a significant degree of diversity\cite{baud2012mutual}.
However, greater variations introduce higher risks and costs. 
Therefore, there are always specific reference systems that remain consistent \cite{albers2018reference}, which can be utilized as prior knowledge.
Our prior knowledge indicates that the bolts are located only in specific regions, oriented along the motor's axis. We can simplify the initially challenging task of 6D pose estimation for small objects that lack distinct features by focusing on the prominent features of the motor.
Additionally, estimating high-precision 6D pose solely from a single model is still challenging. Therefore, we attempted to decompose it into multiple, more precise steps using more comprehensive information sources.

The remainder of this paper is structured as follows: 
Section \ref{sec:relatedWork} summarizes related works. 
Section \ref{sec:Methodology} describes the technical details of our bolt 6D pose estimation pipeline. 
Section \ref{sec:Experimental} provides experimental results, while section \ref{sec:Conlusion} gives a conclusion. 

\section{Related Work}
\label{sec:relatedWork}
\textbf{6D Pose Estimation with RGB(D) Data.}
Initially, methods based on single RGB images were applied, These methods extract distinctive invariant local features from the images and subsequently match them with their corresponding features in 3D space \cite{lowe1999object, lowe2004distinctive,  peng2019pvnet}.
Based on the 2D-3D correspondences, algorithms, such as PnP \cite{fischler1981random} and ICP \cite{besl1992method}, are applied to estimate the 6D pose.
However, the application of RGB-based methods is often an issue when dealing with occlusions or objects without significant texture and discriminative color.

Therefore, additional depth information has been introduced.
MV3D \cite{chen2017multi}, Frustum PointNets \cite{qi2018frustum} and AVOD \cite{ku2018joint} apply convolutional neural networks (CNN) to extract features from RGB images, LIDAR Bird view and LIDAR Front view.
Subsequent to information fusion, another CNN outputs the positions of bounding boxes, with the assumption that they are all situated on a common plane.
PointFusion \cite{xu2018pointfusion} and DenseFusion \cite{wang2019densefusion} employ PointNet \cite{qi2017pointnet} as the feature extractor for point cloud data and CNN as the feature extractor for RGB images. The information from both sources is combined together by a Dense Fusion module.
This method evolves into MV6D \cite{duffhauss2022mv6d} by employing multi-view point clouds and introducing the detection of 3D center points in addition to key points.

\begin{figure*}[t]
\centering
\subfigure[Source and Target]{\includegraphics[width=0.135\linewidth, trim=0 20 0 5, clip]{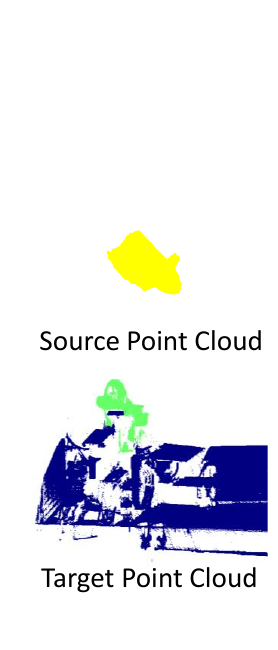}\label{fig: Source and Target point cloud}}
\subfigure[Coarse registration]{\includegraphics[width=0.279\linewidth, trim=0 20 0 5, clip]{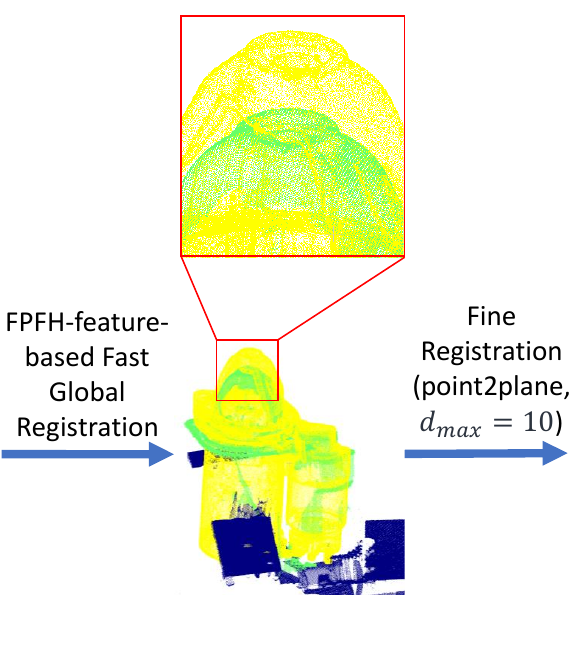}\label{fig: Global Registration}}
\subfigure[Fine registration]{\includegraphics[width=0.535\linewidth, trim=0 20 0 5, clip]{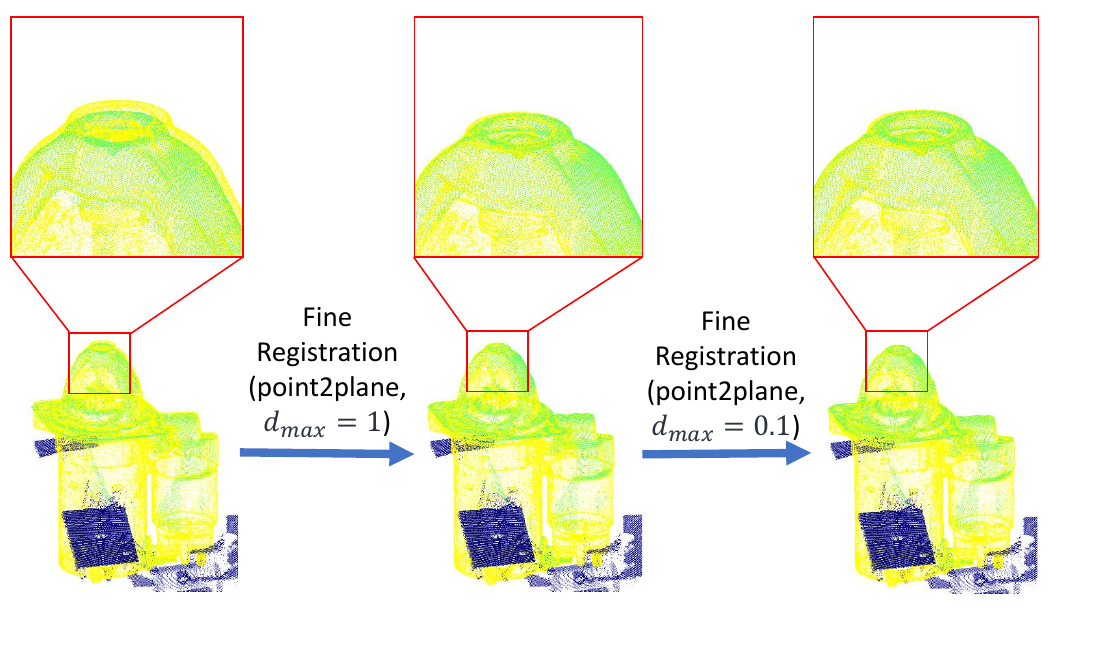}\label{fig: Fine Registration}}
\vspace{-0.2cm}
\caption{Registration between the source point cloud (full motor) and the target point cloud (real scene). A coarse registration is first performed, followed by three successive fine registration processes.}
\label{fig: registration}
\end{figure*}

\textbf{6D Pose Estimation with Point Cloud Data.}
With the rapid technological advancements in LiDAR sensors, which output data in the form of point clouds, a multitude of  point cloud based pose estimation methods have been developed \cite{chen2020survey,fernandes2021point}.
Early research aimed to expand classical convolutional neural networks into voxelized 3D space to form an end-to-end model \cite{li20173d}. However, 3D convolution is prohibitively computationally expensive for practical applications. Therefore,
more recent approaches have employed various other algorithms, such as PointNet \cite{qi2017pointnet}, PointNet++ \cite{qi2017pointnet++}, and DGCNN \cite{wang2019dynamic}, to encode geometric features. A deep Hough voting network was developed for end-to-end 3D object detection \cite{qi2019deep}.
The success of Transformer \cite{vaswani2017attention} in 2D images has inspired transformer-based algorithms. Building upon Point Cloud Transformer (PCT) \cite{guo2021pct}, the Multiscale Point Cloud Transformer (MSPCT) \cite{wu2023mpct} was introduced to address point cloud data.

\section{Methodology}
\label{sec:Methodology}
\subsection{Full Pipeline}
We employ three distinct coordinate systems in our study. The first is the robot coordinate system, referred as Coordinate System $\mathcal{A}$, which has its origin at the center of the robot arm's base and serves as the final reference for all the results in the actual remanufacturing process. The second is the Zivid camera coordinate system, denoted as Coordinate System $\mathcal{B}$, which is used for the scanned single-view point clouds. The third one is the coordinate system of the extra input of full motor point clouds, denoted as Coordinate System $\mathcal{C}$. 
The objective of this multi-stage pipeline is to estimate the 6D pose of motor bolts in the robot's coordinate system, orientation $_{}^{\mathcal{A}}R_{bolt}^{} \in SO (3)$ and position $_{}^{\mathcal{A}}\textbf{t}_{bolt}^{} \in \mathbb{R}^3$, with the additional help of certain extra input data and prior knowledge of the starter motor. 
As illustrated in Fig.\ref{fig:pipeline}, our full pipeline consists of five stages: (i) binary segmentation on real-world scene point cloud; (ii) part segmentation on the extra input of full motor point cloud; (iii) registration between two point clouds; (iv) bolt point clustering for bolt position; and (v) normal alignment for bolt orientation.

\begin{figure}
\centering
\subfigure[]
{\includegraphics[width=0.22\linewidth, trim=0 0 0 0, clip]{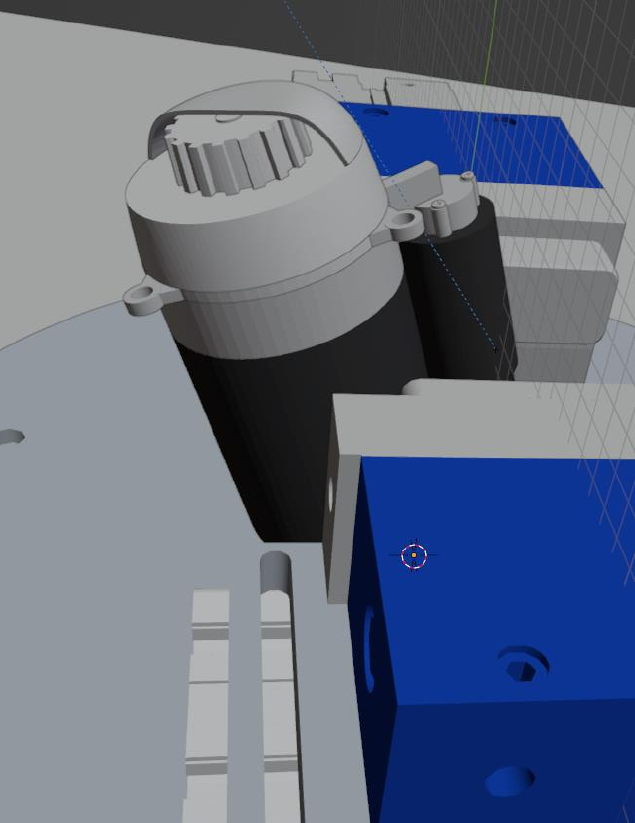}\label{fig: Model for Binary Segmentation}}
\subfigure[]
{\includegraphics[width=0.22\linewidth, trim=0 0 0 0, clip]{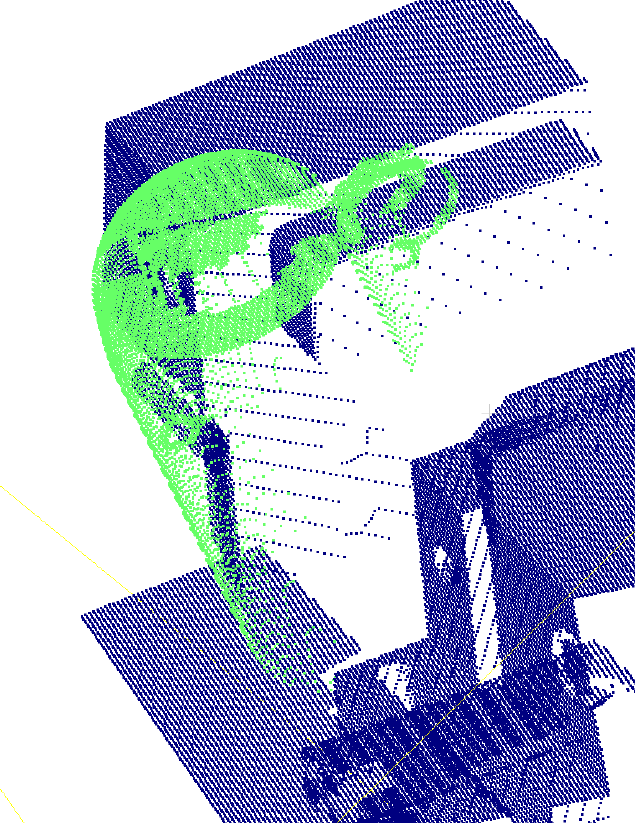}\label{fig: Point Cloud for Binary Segmentation}}
\subfigure[]
{\includegraphics[width=0.22\linewidth, trim=0 0 0 0, clip]{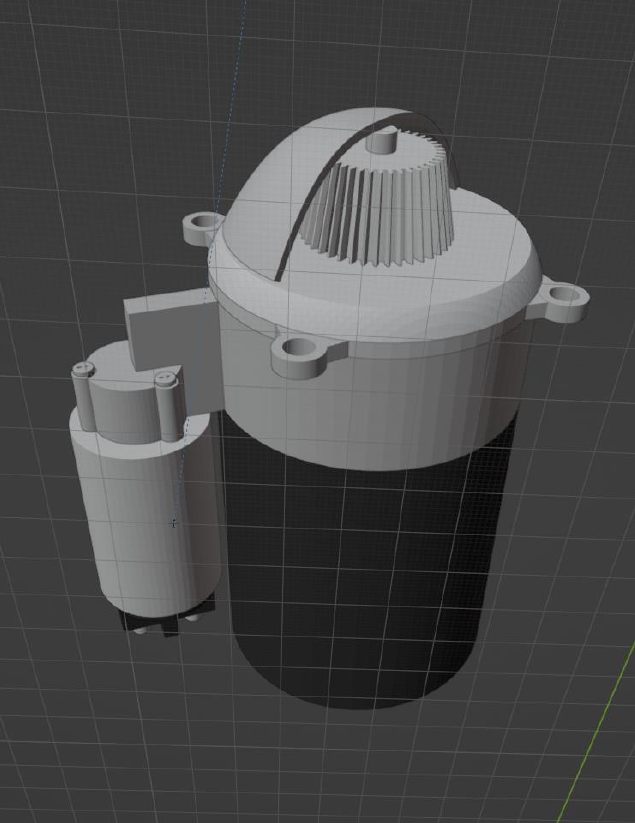}\label{fig: Model for Part Segmentation}}
\subfigure[]
{\includegraphics[width=0.22\linewidth, trim=0 0 0 0, clip]{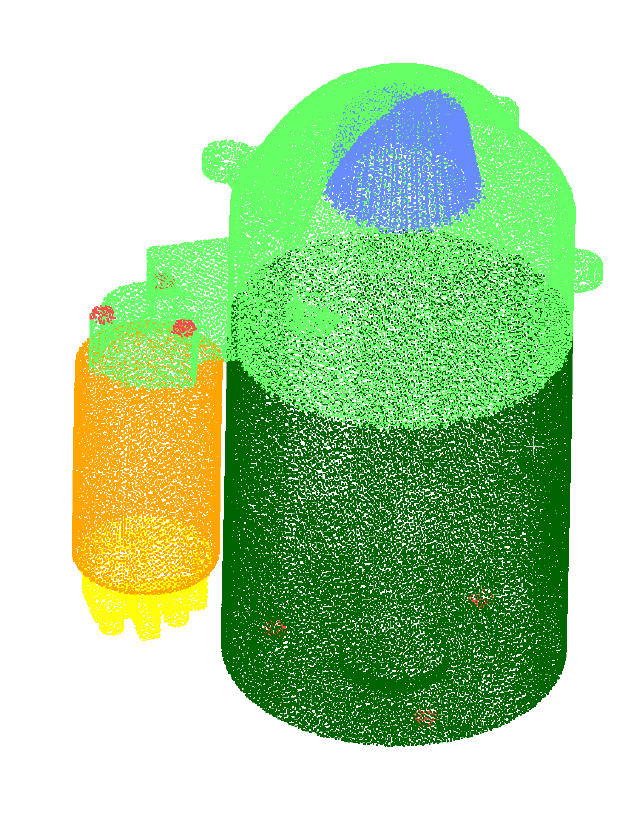}\label{fig: Point Cloud for Part Segmentation}}
\vspace{-0.2cm}
\caption{Synthetic datasets created with Blender. (a, b) Synthetic scene and automatically labeled point cloud for binary segmentation. (c, d) Synthetic motor and automatically labeled point cloud for part segmentation. \vspace{-0.4cm}}
\label{fig: Synthesis Dataset}
\end{figure}

The full model point cloud of the starter motor, which can be generated from its CAD model or obtained through a scanner, is acquired as prior knowledge in this study. The extra input full model point cloud is first fed into a deep learning model that segments it into 6 semantic parts.
The orientation of bolt $_{}^{\mathcal{C}}R_{bolt}^{}$ is determined based on the orientation of the full motor $_{}^{\mathcal{C}}R_{motor}^{}$. Prior knowledge suggests that the orientation of the bolt $_{}^{\mathcal{C}}R_{bolt}^{}$ can either be the same as or opposite to $_{}^{\mathcal{C}}R_{motor}^{}$.
\vspace{-0.5cm}
\begin{equation}\label{eq:motor to bolt}
_{}^{\mathcal{C}}R_{bolt}^{} =
\begin{cases}
\ \ \ _{}^{\mathcal{C}}R_{motor}^{} & \text{, bolt on the top} \\
-\ _{}^{\mathcal{C}}R_{motor}^{} & \text{, bolt on the bottom} 
\end{cases}
\vspace{-0.4cm}
\end{equation}
However, the full motor point clouds are typically from different sources or scanned in different positions, which means $_{}^{\mathcal{C}}R_{motor}^{}$ is unknown.
Therefore, a segmentation result-based normal alignment method is proposed, allowing us to align the motors to obtain their respective $_{}^{\mathcal{C}}R_{motor}^{}$ and determine $_{}^{\mathcal{C}}R_{bolt}^{}$ accordingly.
The 3D positions of the bolts $_{}^{\mathcal{C}}\textbf{t}_{bolt}^{}$ within coordinate system $\mathcal{C}$ is represented by the clustered centers of individual bolt point groups in the segmented full motor point cloud.

During remanufacturing, a single-view point cloud is captured by a Zivid camera.
The single-view point cloud is passed through a deep learning model, which distinguishes between the clamping system and the motor. Using the point cloud of this motor as the target and the full motor point cloud as the source, a registration process is executed, resulting in a transformation matrix from the extra input coordinate system to the Zivid coordinate system $T_{\mathcal{C}\to \mathcal{B}}$.
As a result, the 6D information of the bolts in the Zivid coordinate system and the robot's coordinate system can be obtained from following equations:
\vspace{-0.4cm}
\begin{equation}\label{eq:full model to zivid}
\begin{bmatrix}
 _{}^{\mathcal{B}}R_{bolt}^{} & _{}^{\mathcal{B}}\textbf{t}_{bolt}^{}\\
 0 & 1
\end{bmatrix}\
=\ T_{\mathcal{C}\to \mathcal{B}}\
\begin{bmatrix}
 _{}^{\mathcal{C}}R_{bolt}^{} & _{}^{\mathcal{C}}\textbf{t}_{bolt}^{}\\
 0 & 1
\end{bmatrix}
\end{equation}
\vspace{-1.2cm}
\begin{equation}\label{eq:zivid to robot}
\begin{bmatrix}
 _{}^{\mathcal{A}}R_{bolt}^{} & _{}^{\mathcal{A}}\textbf{t}_{bolt}^{}\\
 0 & 1
\end{bmatrix}\
=\ T_{\mathcal{B}\to \mathcal{A}}\
\begin{bmatrix}
 _{}^{\mathcal{B}}R_{bolt}^{} & _{}^{\mathcal{B}}{\textbf{t}}_{bolt}^{}\\
 0 & 1
\end{bmatrix}
\end{equation}
\vspace{-0.9cm}

\begin{figure*}[t]
\centering
\subfigure[xoy-plane Determination]{\includegraphics[width=0.461\linewidth, trim=0 20 0 18, clip]{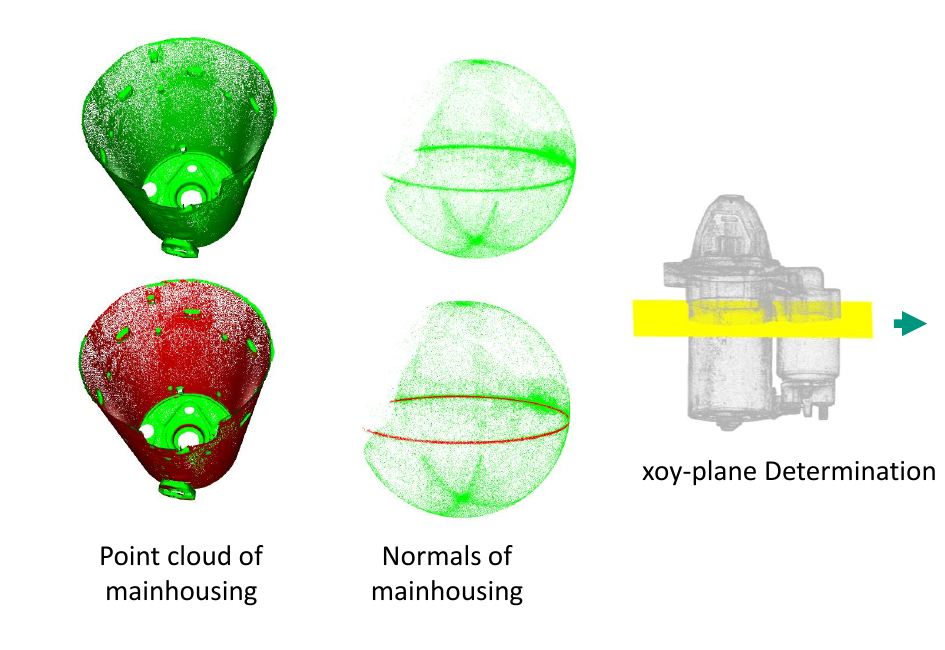}\label{fig: xoy-plane Determination}}
\subfigure[x,y,z-axis Determination]{\includegraphics[width=0.489\linewidth, trim=0 20 0 18, clip]{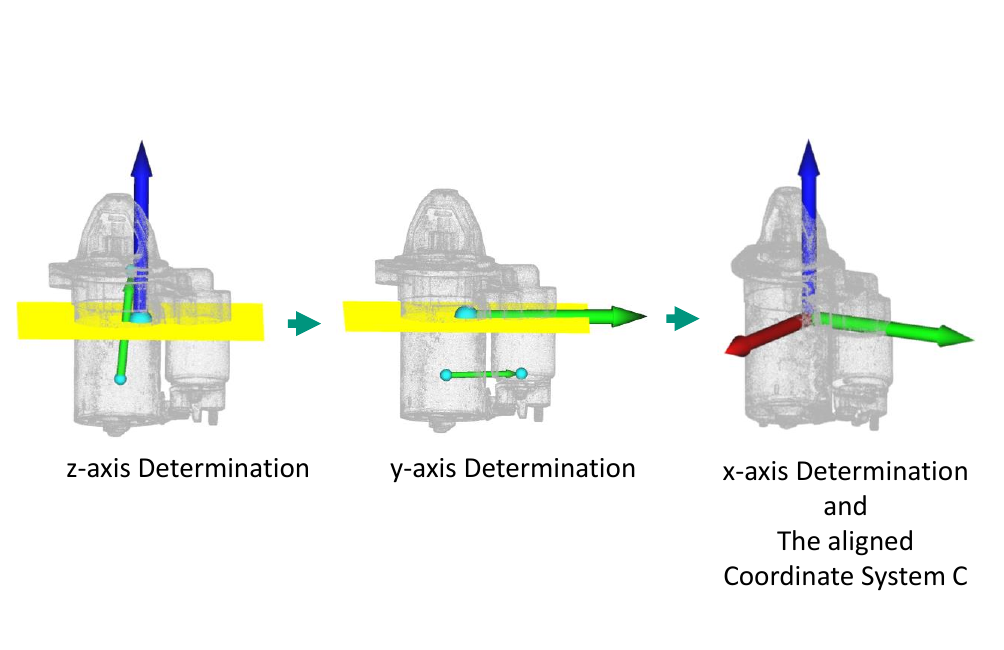}\label{fig: x,y,z-axis Determination}}
\vspace{-0.2cm}
\caption{Normal alignment for motor rotation computation. Three axes directions are computed one by one with the help of part segmentation results.}
\label{fig:alignment}
\end{figure*}

\subsection{Scene Binary Segmentation and Motor Part Segmentation}\label{sec:segmentation}
For this study, we have access to only 18 starter motors from a series. To obtain a full model data set, we utilize a T-scanner to scan the motors. Subsequently, the motors are secured in a clamping system in accordance with the remanufacturing process requirement. They are scanned by a fixed Zivid camera. And a real-world single-view dataset is obtained. However, these data sets proved insufficient in terms of size and representativeness for our two segmentation tasks. The allocation and annotation of significantly larger data sets are both costly and time-consuming, rendering it infeasible. As an alternative, we turn to synthetic data sets generated by a Blender add-on called MotorFactory \cite{wu2022motorfactory}. This add-on initially creates a substantial number of motors with diverse specification, as shown in Fig. \ref{fig: Model for Binary Segmentation} and Fig. \ref{fig: Point Cloud for Binary Segmentation}, outputs them as synthetic single-view point clouds with clamping system from various viewpoints and as synthetic full model point clouds without clamping system as shown in Fig. \ref{fig: Model for Part Segmentation} and Fig. \ref{fig: Point Cloud for Part Segmentation}.

Binary segmentation and motor part segmentation are carried out using two structurally identical deep learning neural networks, both of which are based on PCT \cite{guo2021pct}.
The PCT network for binary segmentation is initially pretrained on the synthetic single-view point cloud data set and then fine-tuned on the real-world single-view point cloud data set. Similarly, the PCT network for part segmentation is pretrained and fine-tuned on their corresponding synthetic and real-world data set.

\subsection{Registration}
To determine the 6D pose of bolts in the motor in clamping system, $_{}^{\mathcal{B}}R_{bolt}^{}$ and $_{}^{\mathcal{B}}\textbf{t}_{bolt}^{}$, a Transform matrix, $T_{\mathcal{C}\to \mathcal{B}}$ is required. As the full motor, considered as source point cloud, coincide after registration with the motor in clamping system, regarded as target point cloud, the bolts in the source will also coincide with bolts in the target. The transformation matrix obtained during this registration process is the desired $T_{\mathcal{C}\to \mathcal{B}}$. Since errors can result in a detrimental impact on instruments in subsequent process, such as the screwdriver bit on the robot arm, a high level of precision is demanded in our results. 
Simultaneously, a significant 6D pose difference exists between our source and target, making it challenging to establish correct matches for fine registration methods.  Therefore, we divided our approach into two steps: coarse registration and fine registration.

In the coarse registration step, we downsample the point cloud, estimate normals, and compute a 33-dimensional Fast Point Feature Histogram (FPFH) \cite{rusu2009fast} for each point. The FPFH feature comprehensively describes the local geometric properties of a point. These features are then employed in Fast Global Registration \cite{zhou2016fast}, the results of which are depicted in Fig. \ref{fig: Global Registration}.

For the fine registration step, we employ Iterative Closest Point (ICP) \cite{besl1992method}. As we reduce the maximum matching distance $d_{max}$, the likelihood of false matches decreases, yielding a smaller registration error. However, it is vital to note that the initial 6D poses of the source and target point clouds must be close, or an insufficient number of matches may result in registration failure. To overcome this limitation, we divide the fine registration process into multiple steps, with the results of each step serving as the source point cloud for the subsequent one, and with a decreasing maximum matching distance $d_{max}$. 
As demonstrated in the Fig. \ref{fig: Fine Registration}, we implement a three-steps fine registration approach. The rationale behind our choice of this approach, along with its details, is explained in the section \ref{sec:Experimental}.This iterative approach results in a precise registration, where the registered full motor point cloud perfectly overlaps with the point cloud of the motor in segmented point cloud.

\subsection{Bolt Point Clustering}
After part segmentation described in section \ref{sec:segmentation}, the point cloud is segmented into bolt and five other classes. Given our prior knowledge of the motor structure, points belonging to the same bolt should be grouped together, while those from different bolts should be separated. However, the precise number of bolts on the scanned motor is unknown. Additionally, there exists a certain level of noise, and the shape and size of the bolt point clouds can exhibit variations. These characteristics make the Density-Based Spatial Clustering of Applications with Noise (DBSCAN) \cite{ester1996density} exceptionally well-suited for this task. We utilize the DBSCAN algorithm to cluster the bolt points into distinct bolt point clouds, with the centers of these bolt point clouds representing the desired 3D positions $_{}^{\mathcal{C}}\textbf{t}_{bolt}^{}$.

\begin{table}
\centering
\caption{Different registration methods.}
\label{table:Registration}
\vspace{2pt}
\resizebox{1\linewidth}{!}{
\begin{tabular}{cc|cc}
\toprule
Method & $d_{max}$ (mm) & RMSE (mm) & Process time (s) \\ \midrule
\multirow{3}{*}{point2point} & 10 & 4.11 & 9.64 \\
 & 10 $\rightarrow$ 1 & 4.27e-1 & 13.82 \\
 & 10 $\rightarrow$ 1 $\rightarrow$ 0.1 & 7.66e-2 & 21.93 \\ \midrule
\multirow{3}{*}{point2plane} & 10 & 7.96e-1 & \textbf{3.72} \\
 & 10 $\rightarrow$ 1 & 3.44e-1 & 4.65 \\
 & 10 $\rightarrow$ 1 $\rightarrow$ 0.1 & \textbf{7.62e-2} & 6.39 \\ \bottomrule
\end{tabular}}
\end{table}

\begin{figure}
\centering
\includegraphics[width=0.9\linewidth, trim=0 0 0 0, clip]{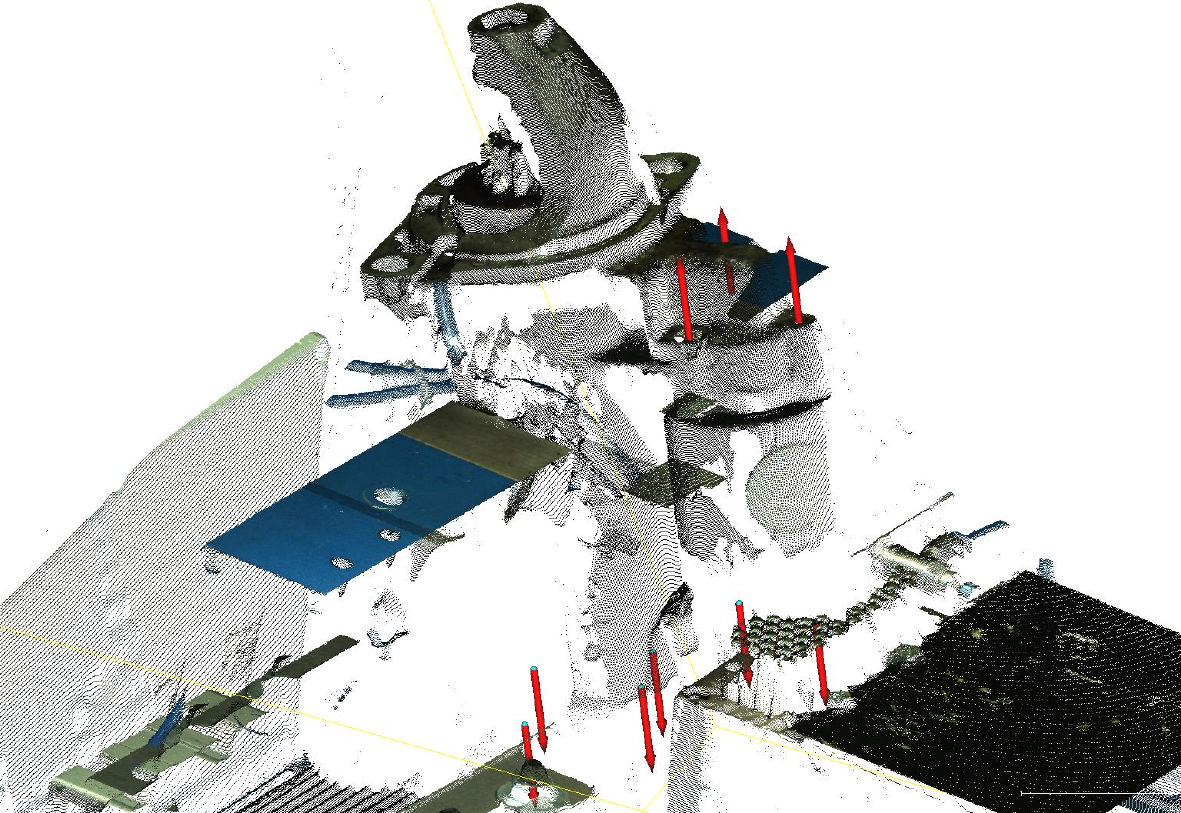}
\caption{6D pose of bolts in a real-world case. Note that the red arrows are the real numerical results computed with our proposed pipeline.}
\label{fig: demo}
\end{figure}

\subsection{Normal Alignment}\label{sec:normal alignment}
The alignment of the full motors is as follows: The center point of the motor coincides with the origin of the coordinate system. The z-axis is parallel to the cylindrical axis of a motor part known as the Main Housing. The positive direction of the z-axis points from the Main Housing part towards the Connector part. The y-axis is positioned perpendicular to the z-axis, and it is positioned such that the center of the Solenoid part lies in the yoz plane.
Based on prior knowledge, the orientation of the bolts $_{}^{\mathcal{C}}R_{bolt}^{}$ can be determined by the orientation of the aligned motor $_{}^{\mathcal{C}}R_{motor}^{}$, as represented in Equation \ref{eq:motor to bolt}.

According to the coordinate system definition, the normal vectors of points on the main housing cylinder are parallel to the xoy plane. Based on the segmented full motor point cloud, we obtain the normal vectors for each point on the Main Housing part \cite{hoppe1992surface}. These normal vectors are represented as points in space, as shown in the Fig. \ref{fig: xoy-plane Determination}. It can be observed that these points form a spherical surface, which is most dense near the equator of the sphere. This equatorial plane corresponds to the motor's xoy plane and is determined by RANSAC \cite{fischler1981random}.

With this plane as a reference, other orientations can be determined as Fig. \ref{fig: x,y,z-axis Determination}. The z-axis should be perpendicular to the xoy plane, directing from the Main Housing to the Connector. The positive direction of the y-axis is determined by projecting the vector from the center of the Main Housing to the center of the Solenoid onto the xoy plane. The x-axis is determined using the z-axis and y-axis with the right-hand rule. The motor's orientation $_{}^{\mathcal{C}}R_{motor}^{}$ aligns with the positive z-axis direction.

\begin{table}
\centering
\caption{Variations on backbones of deep learning architecture.}
\label{table:backbone}
\vspace{2pt}
\resizebox{0.8\linewidth}{!}{
\begin{tabular}{cccc}
\toprule
Model & \begin{tabular}[c]{@{}c@{}}Overall\\ Accuracy (\%)\end{tabular} & mIoU (\%) & \begin{tabular}[c]{@{}c@{}}Bolt\\ IoU (\%)\end{tabular} \\ \midrule                            
PointNet & 66.82 & 59.63 & 26.10 \\
PointNet++ & 80.64 & 72.13 & 48.16 \\
DGCNN & 84.57 & 75.07 & 48.04 \\
PCT & \textbf{90.47} & \textbf{84.58} & \textbf{64.92} \\ \bottomrule 
\end{tabular}}
\end{table}

\begin{table} 
\centering
\caption{Variations on Settings of deep learning architecture.}
\label{table:supper_parameter}
\vspace{2pt}
\resizebox{1\linewidth}{!}{
\begin{tabular}{ccc|ccc}
\toprule
\begin{tabular}[c]{@{}c@{}}Point\\ Number\end{tabular} & Pretrain & Warm\_up & \begin{tabular}[c]{@{}c@{}}Overall \\ Accuracy\\ (\%)\end{tabular} & \begin{tabular}[c]{@{}c@{}}mIoU\\ (\%)\end{tabular} & \begin{tabular}[c]{@{}c@{}}Bolt \\ IoU\\ (\%)\end{tabular} \\ \midrule
2048 & - & - & 79.35  & 72.80 & 44.35\\
2048 & - & \checkmark & 83.02  &76.07  & 44.80\\
2048 & \checkmark & - & 87.58 & 81.12& 59.45 \\
2048 & \checkmark & \checkmark & \textbf{90.47} & \textbf{84.58} & \textbf{64.92} \\
1024 & \checkmark & \checkmark & 86.21 & 79.58 & 53.85 \\
4096 & \checkmark & \checkmark &71.89  & 61.33 & 33.20 \\ 
4096 (x2 epoch) & \checkmark & \checkmark & 83.49 &76.72  &  55.43\\ \bottomrule 
\end{tabular}}
\end{table}

\section{Experimental Results}\label{sec:Experimental}

\textbf{Registration:}
We conduct experiments on two ICP variants: one based on point-to-point matches between the source and target, and another relying on matches between points in the source and planes in the target, which are determined by the points and their normals. Additionally, we test several combinations of maximum matching distances. The results are presented in Table \ref{table:Registration}, which demonstrates that as the number of fine registration steps increases, processing time exhibits linear growth, and registration accuracy improves significantly. Furthermore, compared with the point-to-point algorithm, the point-to-plane algorithm exhibits notably reduced processing time. In this study, although real-time performance holds importance, we prioritize precision over real-time performance, leading us to choose a point-to-plane registration approach with three fine-registration steps.

\textbf{Full Model Segmentation:}
Our primary goal is to identify the most suitable combination of structures and algorithms. 
Given that our test set comprises only 18 motors, which is extremely small, the results derived from a single train-test split setting lack representativeness. To mitigate this issue, we use a method of cross-validation. In each training, 2 random motors are selected as the test set, and the other 16 motors are used as the training set. We perform such trainings 5 times and report their mean results as the final cross-validated performance.

Four widely-used backbones are evaluated in our experiments, as depicted in Table \ref{table:backbone}. We find that the PCT \cite{guo2021pct} yields the most favorable results.
Following this, we conducted tests on various hyperparameters based on the PCT, and the most significant experimental results are presented in Table  \ref{table:supper_parameter}. Note that all the experiments in this table use the aforementioned cross-validation. Moreover, for a fairer comparison, we make the 5 train-test splits identical for all the experiments.
From it, we can observe that the effectiveness of performing pre-training on the synthetic dataset is evident.
To mitigate model instability resulting from an excessively high initial learning rate during transfer learning, using a warm-up strategy can further optimize the performance of our network model.
Regarding the input sub-point cloud size for the training, as indicated in the table \ref{table:supper_parameter}, a sampling size of 2048 points performs the best, even better than a sampling size of 4096 points. Since increased sampling size leads to a reduction of the actual interaction steps in one epoch, an additional experiment of 4096 points with doubled training epochs is performed. However, its performance still does not surpass that of the one with 2048 points.

\textbf{Binary Segmentation:}
Binary segmentation is performed on our single-view data set, employing the identical neural network architecture and hyperparameter settings, with the only modification being made to the output fully connected layer to accommodate a reduced number of classes.
And the overall accuracy and mIoU yield the values of 99.80$\%$, and 99.07$\%$.
Given the algorithm's already high accuracy and the inherent limitations related to labeling errors that cannot be entirely eliminated, further tests are deemed unnecessary.

\textbf{Full Pipeline Demo:}
The Fig. \ref{fig: demo} illustrates an example input point cloud in a real-world case. The red arrows depict the computed 6D pose of all 9 bolts.
As shown in Fig. \ref{fig: demo}, our method can accurately determine the 6D pose of bolts, even in cases where they have not been precisely scanned or are occluded.

\section{Conclusion}\label{sec:Conlusion}
This study presents a novel pipeline designed for the precise determination of the 6D pose of bolts leveraging prior knowledge within a remanufacturing line. 
By utilizing a pre-acquired model, we perform part segmentation, normal alignment, and bolt point clustering to extract valuable information, which includes bolt 6D pose on the full motor and the full motor point cloud. 
This extracted knowledge is subsequently transferred to the motors being processing through binary segmentation and registration steps, so that the 6D pose of bolts on the processed motors is determined.
The experimental results illustrate the effectiveness of our pipeline even in challenging case characterized by low-quality scans and occlusion.

\section*{Acknowledgements}
The AgiProbot project is funded by the Carl-Zeiss Foundation.

\bibliography{reference}
\bibliographystyle{unsrt2authabbrvpp}

\end{document}